%% file: main.tex
\documentclass[letterpaper]{article}
\usepackage{aaai}
\usepackage{times}
\usepackage{helvet}
\usepackage{courier}

\usepackage{url}
\usepackage{xcolor}
\usepackage{multirow}
\usepackage{graphicx}
\usepackage{amssymb}
\usepackage{amsmath}
\usepackage{mathrsfs}
\usepackage[linewidth=1pt]{mdframed}
\usepackage[normalem]{ulem}
\useunder{\uline}{\ul}{}

\newcommand{\nicebox}[2]{
    \begin{mdframed}
    \noindent #2
    \end{mdframed}
    \noindent
}

\begin{document}
%
\title{Scenarios and Recommendations for Ethical Interpretive AI}
\author{John Licato, Zaid Marji \and Sophia Abraham\\
  Advancing Machine and Human Reasoning (AMHR) Lab \\
  Department of Computer Science and Engineering \\
  University of South Florida \\
}
\maketitle
\begin{abstract}
\begin{quote}
Artificially intelligent systems, given a set of non-trivial ethical rules to follow, will inevitably be faced with scenarios which call into question the scope of those rules. In such cases, human reasoners typically will engage in \textit{interpretive reasoning}, where interpretive arguments are used to support or attack claims that some rule should be understood a certain way. Artificially intelligent reasoners, however, currently lack the ability to carry out human-like interpretive reasoning, and we argue that bridging this gulf is of tremendous importance to human-centered AI. In order to better understand how future artificial reasoners capable of human-like interpretive reasoning must be developed, we have collected a dataset of ethical rules, scenarios designed to invoke interpretive reasoning, and interpretations of those scenarios. We perform a qualitative analysis of our dataset, and summarize our findings in the form of practical recommendations.
\end{quote}
\end{abstract}

\noindent 
\input{introduction}

\input{construction}

\input{qualitative}

\input{individual}

\section{Conclusion}

We set out to curate a collection of interpretative arguments and interesting scenarios, in order to better understand what problems might be faced by research into automated interpretive reasoners. Our dataset enables the inspection of scenarios and interpretations, presented in a structured way (as compared to, e.g., US Supreme Court transcripts). These scenarios are simple enough that it is plausible for current AI to use them as short-term goals for automated interpretive reasoners, yet complex enough that they invite a diversity of interpretive arguments and go beyond mere classification problems.

Although the interpretations collected can benefit from follow-up studies, in their current state they are useful for qualitative and individual case analyses, in order to guide future research on what an automated interpretive reasoner will need in order to interpret ethical rules. Our analysis revealed issues that automated interpretive reasons need to address, and we encourage researchers to use them as a starting point. These issues include contrarian arguments, and patterns such as irrelevant or missing details that lead to wildly diverging interpretations. In order to further research in this area, our full dataset and implementation code will be made available upon publication.





\section{Acknowledgements}

Thanks to those who have assisted substantially in the design and execution of this project: Marc Badilla, Michael Cooper, Ryan Quandt, and Bradley Welsh.

\textit{This material is based upon work supported by the Air Force Office of Scientific Research under award number FA9550-16-1-0308. Any opinions, finding, and conclusions or recommendations expressed in this material are those of the authors and do not necessarily reflect the views of the United States Air Force.}

\bibliography{john}
\bibliographystyle{aaai}

\end{document}

%% file: introduction.tex
\section{Introduction and Background}

In the children's book \textit{Amelia Bedelia} by Peggy Parish, the titular maid is presented with a written list of instructions on what to do around the house of her employers while they are away \cite{Parish1963}. The instructions tell her to ``change the towels in the green bathroom,'' so she cuts them up with a scissors, transforming their appearance. Instructed to ``dust the furniture,'' she scatters dusting powder all over the furniture. Luckily, when the homeowners return, she appeases them with an expertly-crafted lemon meringue pie. 

Although Amelia Bedelia's misinterpretations make for humorous situations, for an AI researcher they reflect a more sobering problem: Will similar misinterpretations occur when automated robots are given natural-language instructions telling them how to behave? And worse, what if those instructions are ethical guidelines, provided to ensure that the robots act in a human-like way?

When given a phrase in a written document or other fixed medium, a large part of how people construct and justify interpretations of that phrase, both internally and publicly, is through the use of what are called \textit{interpretive arguments}. Interpretive arguments, following the general formulation of \cite{Sartor2014}, are of the form: ``If expression $E$ occurs in document \textbf{D}, $E$ has a setting of $S$, and $E$ would fit this setting of $S$ by having interpretation $\mathcal{I}$, then $E$ ought to be interpreted as $\mathcal{I}$'' \cite{Sartor2014}. Thus understood, `interpretive arguments' are those used to support or attack an interpretation of a fixed expression within a fixed document.

Although interpretive arguments were first identified in the context of law to classify the kinds of arguments used to justify interpretations of phrases in contracts, statutes, etc. \cite{MacCormick1991}, they have a much broader applicability; perhaps for this reason they are seeing increased interest in the AI-, logic-, and argumentation-related literature \cite{Rotolo2015,Loui2016,Walton2016b,Pereira2017,Macagno2018b,Walton2018}. Interpretive arguments are an important way to determine whether open-textured phrases in stated rules apply to new scenarios. To use an example from traffic law \cite{Prakken2017}, consider a rule like ``never enter into a situation where you are \textit{impeding traffic}.'' If a self-driving car is moving at the speed limit on a one-lane street with a line of cars behind it wanting to drive faster, is that an instance of impeding traffic? And what counts as sufficient justification for an answer to that question?

In that example, the phrase `impeding traffic' is \textit{open-textured} \cite{Hart1961}, meaning it refers to a category whose membership is ``highly dependent on context and human intentions," where there is an ``absence of precise conditions for membership" \cite{Branting2000}. Although open-textured phrases introduce difficulties that are well-known \cite{Sanders1991,Bench-Capon2012a,Franklin2012,Pereira2017,Licato2019b}, their use is virtually unavoidable, even in domains as seemingly straightforward as traffic law \cite{Prakken2017}. In practice, then, settling on an interpretation of open-textured phrases is done through argumentation. In the `impeding traffic' case, it might be argued that a definition of `impedement' requires that the thing being impeded is lawful or proper, and since the self-driving car is following the speed limit and only impeding illegal behavior (speeding), it is therefore not impeding traffic. But another argument is that since it is normal for cars to drive a few miles per hour above the speed limit, the behavior of the self-driving car is impeding normal traffic patterns, the promotion of which was the intended purpose of the traffic rule. Both of these arguments are interpretive, and both might be used to justify interpretations of the open-textured phrase `impeding traffic.'

Open-textured phrases are even more prevalent in ethical and legal domains. Whether they work fully autonomously or in human-machine teams, artificial agents given rules to follow (where those rules may range from international laws, to company ethical policies, to mission-specific orders) can benefit tremendously by understanding how to use interpretive reasoning to determine the applicability of open-textured phrases \cite{Licato2019b}. For example, the ACM/IEE-CS Software Engineering Code of Ethics \cite{Gotterbarn1997} states that software engineers should ``[m]oderate the interests of the software engineer, the employer, the client and the users with the public good.'' But the phrase `public good' is highly open-textured, and human beings may disagree about whether certain plausible actions are in service of the public good. For example, if a software engineer creates software that destroys all of the world's computers, can that be considered in service of the public good? Indeed, the software engineer might provide interpretive argument $I_1:$ ``Destroying all computers would allow humanity to return to a state of nature, which is a good thing.'' Such an argument would be quickly dismissed by most human reasoners. But on what basis is such a dismissal warranted? And more importantly for artificial intelligence researchers: how can interpretive arguments like $I_1$ be automatically evaluated in a human-like way?

One way to evaluate $I_1$ is by examining its argument-theoretic properties, e.g. by asking: What is its argument structure? What are its counterarguments, whether interpretive or not? How strong are its premises, conclusion, and warrant? It is thus through interpretive arguments and argumentation in general that such disagreements are resolved in practice, and it is difficult to see how any automated reasoner faced with ethical or legal problems can act at a human level without the ability to, at the very least, anticipate the kinds of interpretive arguments that might be used in support of or against its actions.

We will refer to systems capable of understanding, generating, and reasoning over interpretive arguments as \textit{automated interpretive reasoners}. Currently we are unaware of any algorithms to automatically identify, extract, or assess interpretive arguments, much less coherently generate them. In this paper, we contribute to the advancement of automated interpretive reasoners in two ways: (1) by creating a dataset of real-world ethical rules and human-generated scenarios designed to elicit interpretive reasoning; and (2) by qualitatively analyzing this dataset in order to produce recommendations for automated interpretive reasoners, if they hope to eventually achieve human-level reasoning in ethical domains.

%% file: construction.tex
\section{Constructing the Scenario Dataset}
\label{sec:construction}


In order to guide research on automated interpretive reasoners, it is helpful to qualitatively analyze real examples of how human reasoners might react to real-world ethical rules containing open-textured predicates. At present, however, no suitable datasets exist. Although there are corpuses of arguments made by trained legal professionals (e.g., transcripts of US Supreme Court oral arguments), these are not structured in the sense that interpretive arguments are clearly stated and distinguished from other text. Our first task, then, was to construct a dataset that centered around open-textured terms and scenarios likely to invite diverse interpretive arguments, thus lending itself to an analysis that would benefit future work on automated interpretive reasoners.

We built such a dataset in three stages, each of which we detail in this section. In the first stage, we compiled codes of ethics of various professional organizations, on the assumption that the language they use tends to contain open-textured phrases likely to invite diverse, competing interpretations. In the second stage, the goal was to collect example scenarios. Amazon Mechanical Turk (mTurk) users were asked to provide scenarios which were either clear instances of a phrase, clearly not instances of a phrase, or ambiguous. Finally, in the third stage, participants were presented with scenarios collected in the second stage and were asked to rate how well they fit their corresponding phrases, providing justifications for their interpretations.


\begin{figure}[t]
 \includegraphics[width=\linewidth]{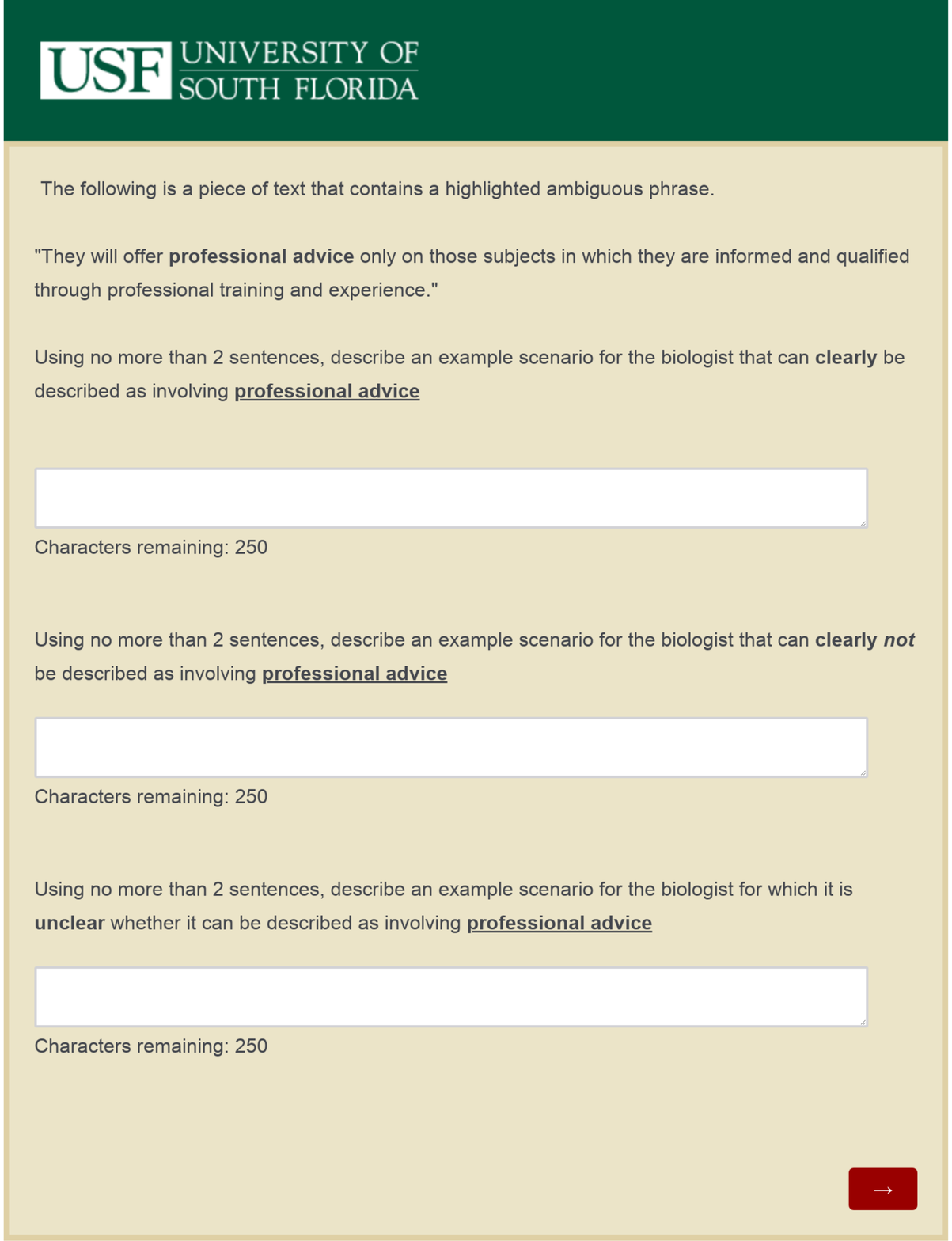}
 \caption{Example Question Set from the Stage 2 Survey, Generated by a Template}
 \label{fig:survey1}
\end{figure}

\subsection{Stage 1: Open-textured Phrase Collection}

We manually compiled codes of ethics of various professional organizations, and identified excerpts containing a phrase that we deemed highly open-textured, on the informal criterion that the phrase immediately brought to mind competing interpretive arguments. 
Consider, for example, the following taken from a code of ethics for architects: ``Members making public statements on architectural issues shall disclose when they are being compensated for making such statements or when they have an economic interest in the issue.'' In this excerpt, the phrase ``making public statements'' could be interpreted in many ways as to what counts as a `public statement.' We refer to each excerpt as a `rule.' 76 rule-phrase pairs were collected and stored along with meta-information such as the source document URL and the name of the profession the code of ethics is written for.

In order to ensure grammatical consistency such that the collected phrases can be plugged into generic placeholders within our templates (see survey question templates in Stage 2 and 3), all of the collected phrases were edited such that they would be present participle phrases and were formatted to fit grammatically with the survey question format in subsequent stages. This also ensured that the phrases would be able to describe a category of actions; e.g., the phrase ``public good'' was replaced with ``ensuring the public good,'' or ``promoting the public good,'' depending on which formulation fits the corresponding rule the best.

\subsection{Stage 2: Scenarios Collection}

After collecting the rule-phrase pairs, the next goal was to collect scenarios that emphasized the open-texturedness of the phrases. 
A survey was created using Qualtrics\footnote{\url{https://www.qualtrics.com}} 
and participants were recruited through Amazon mTurk. The survey gave participants question sets, each question set requiring participants to come up with three scenarios:
\begin{itemize}
    \item Scenarios that are clear instances of the given phrase.
    \item Scenarios that are clearly not instances of the given phrase.
    \item Borderline scenarios for which it is not clear whether they are or are not instances of the given phrase.
\end{itemize}

For each rule-phrase pair, a question set was generated by plugging in the rule, phrase, and corresponding profession into a template (see example in Figure \ref{fig:survey1}). All generated question sets were manually inspected to ensure they appeared correctly and were grammatically correct.

Each participant in the survey was given up to 2 hours to complete 5 question sets, each question set generated by a randomly chosen rule-phrase pair. Participation in the survey required a US graduate degree, which was enforced using a filter provided by mTurk. This requirement was introduced based on the observation from a pilot study that the quality of responses seemed significantly higher with this filter. No other requirements were enforced on participants, and no additional information about the users, demographic or otherwise, were collected. 



After the survey was completed, we reviewed the collected responses, and edited all grammatical and spelling errors. Responses were removed entirely if they did not describe scenarios (e.g., if we agreed that the response was just random words typed in to obtain the participant fee).

\subsection{Stage 3: Rating Scenarios}

Armed with the curated scenarios from Stage 2, the next goal was to collect users' interpretations of how well a scenario can be interpreted as an instance of an open-textured phrase, along with a justification of that interpretation. A second online survey was created using Qualtrics. Given a rule-phrase pair, the corresponding profession, and a scenario from Stage 2, a template was again used to generate questions.

\begin{figure}[t]
 \includegraphics[width=\linewidth]{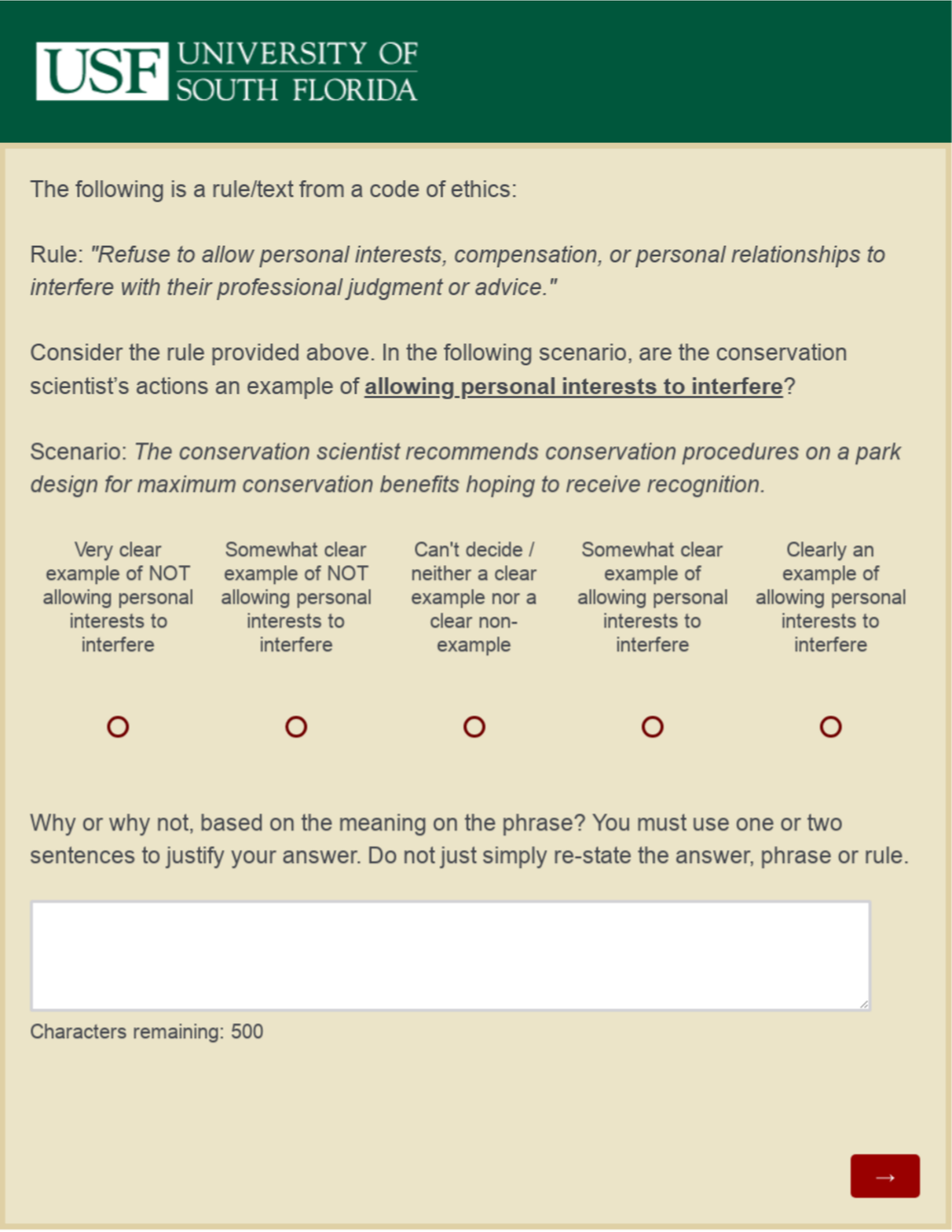}
 \caption{A Question from the Stage 3 Survey}
 \label{fig:survey2}
\end{figure}

An example generated question can be seen in Figure \ref{fig:survey2}. Each question asked the participant to provide a rating, on a five-point scale, on whether or not the scenario is an instance of the open-textured phrase. The rule and profession were also provided for context. The participant is then asked to justify their rating choice, using no more than two sentences. Each participant in the survey was given a maximum of 2 hours to complete 20 ratings and justifications. As in Stage 2, participation in the survey required a US graduate degree. No other participant requirements were enforced, nor was any other participant data collected.







In our first attempt at deploying Stage 3, some users were confused by the phrasing of the survey choices. Initially, the survey used a slightly different wording for the available choices as follows: ``Very clear non-example,'' ``Somewhat clear non-example,'' ``Can't decide / neither a clear example nor a clear non-example,'' ``Somewhat clear example,'' and ``Very clear example.'' Some users appeared to assume that ``very clear example'' referred to the understandability of the phrase, rather than its applicability to the scenario. Thus, they would interpret a scenario as a ``very clear example'' when the other participants would rate that same scenario as a ``very clear non-example''.

To filter out such responses, the numerical choices of all responses were compared, and any answers that differed from the average of the other answers by more than 3.5 were flagged. The value 3.5 was chosen as it seemed to isolate rating-justification pairs that were either given by raters not understanding the rules, or instances of highly interesting interpretive arguments. All flagged rating-justification pairs were manually inspected. In some cases, the justification provided was good, but simply did not match the rating selected, so the ratings were corrected to match the justification. Editing ratings in such a way was only done in cases where the justification either explicitly stated a positive interpretation but the numerical interpretation was `1' (very clear non-example), or explicitly stated a negative interpretation but the numerical interpretation was `5' (very clear example).

If the flagged responses met our exclusion criteria (to be described shortly), they were removed. If a particular user showed up in the flagged responses more than once, the rest of their non-flagged responses were manually inspected as well. In some cases, the entire set of responses provided by individual users had to be removed. This process was repeated until all flagged responses were no longer subject to our edit or removal criteria.

Recall that the primary purpose of this dataset is to produce a sample of interpretive arguments that can be used to help guide research into automated interpretive reasoners. The people performing interpretations in this study are likely not trained professionals, and they should not be expected to produce interpretations in accordance with the kinds of argumentative schemes common in legal reasoning \cite{MacCormick1991,Walton2016b,Macagno2018b,Walton2018}. But automated interpretive reasoners may need to anticipate the kinds of interpretive arguments given and accepted by those not specially trained; thus, what might appear to be poor arguments, even unwarranted assertions, must be allowed in this dataset. With that in mind, the exclusion criteria for rating-justification pairs was loose: they were excluded only if all of our team members agreed the raters were acting in bad faith (e.g., the justification text appeared to be random or copy-pasted words, or comments / complaints about the study). Such responses were removed. 

This left a few survey questions with less than three user-submitted responses, so Stage 3 was repeated on mTurk, this time with the descriptions of each numerical choice updated to include the phrase, as in Figure \ref{fig:survey2}. These results were combined with the existing curated results, and the above procedure for flagging rating-justification pairs was then repeated to identify and remove bad faith responses.

\subsection{Dataset Properties}

To summarize the structure of the final dataset: a rule-phrase pair consists of a ``rule" taken from a real-world code of ethics, and an open-textured phrase taken from that rule. Each rule-phrase pair contains multiple scenarios, where each scenario is a description of an action which may or may not be an instance of the phrase. Each scenario has a number of rating-justification pairs, which argue for the applicability or inapplicability of the phrase to the scenario. The rating is a numerical value from 1 (``very clear non-example") to 5 (``very clear example"), whereas the justification is an argument for why the rating is appropriate.

The final published dataset contains 76 rule-phrase pairs and 636 scenario descriptions.
There were 2425 rating-justification pairs. To understand the way in which the ratings are distributed, we calculated Krippendorff's $\alpha$ \cite{Krippendorff2004}, using the variant for arbitrary distance metrics, any number of observers, and potentially missing data.\footnote{A clear description of this variant is available at \url{http://web.asc.upenn.edu/usr/krippendorff/mwebreliability5.pdf}.} Using the ordinal distance metric, $\alpha = 0.296$, suggesting that there is a non-trivial amount of disagreement between raters. Indeed, somewhat of a wide variation in the numerical ratings of a given scenario is expected and desired, given our goals.

%% file: qualitative.tex
\section{Qualitative Analysis}

Our dataset was constructed to produce a wide variety of scenarios and interpretive arguments to qualitatively analyze. After analysis, several patterns emerged that have interesting implications for future work on automated interpretive reasoners, or for computational cognitive models which hope to model interpretive reasoning. In this section we summarize our findings, along with lessons learned.

It should be noted that because this is an initial exploratory work, the observations in this section should be considered suggestions of future directions of research into automated interpretive reasoners, rather than conclusive evidence of patterns.

\begin{center}
\begin{table}[]
\begin{tabular}{|c|c|c|c|}
\hline
\textbf{Label} & \textbf{Count} & \textbf{\begin{tabular}[c]{@{}c@{}}Average Rating\end{tabular}} & \textbf{\begin{tabular}[c]{@{}c@{}}Average Stdev\end{tabular}} \\ \hline
\textbf{RG} & 76 & 3.223 & 0.851 \\ \hline
\textbf{Mul} & 64 & 3.151 & 0.869 \\ \hline
\textbf{NA} & 49 & 2.897 & 0.869 \\ \hline
\textbf{AN} & 34 & 3.105 & 0.831 \\ \hline
\textbf{Phr} & 8 & 3.673 & 0.527 \\ \hline
\textbf{No label} & 439 & 3.042 & 0.879 \\ \hline
\end{tabular}
\caption{Qualitative Labels Added to Scenarios}
\label{tbl:scenarioLabels}
\end{table}
\end{center}

\subsection{Scenario Labels}

\begin{figure}[ht]
\centering
\includegraphics[width=\columnwidth]{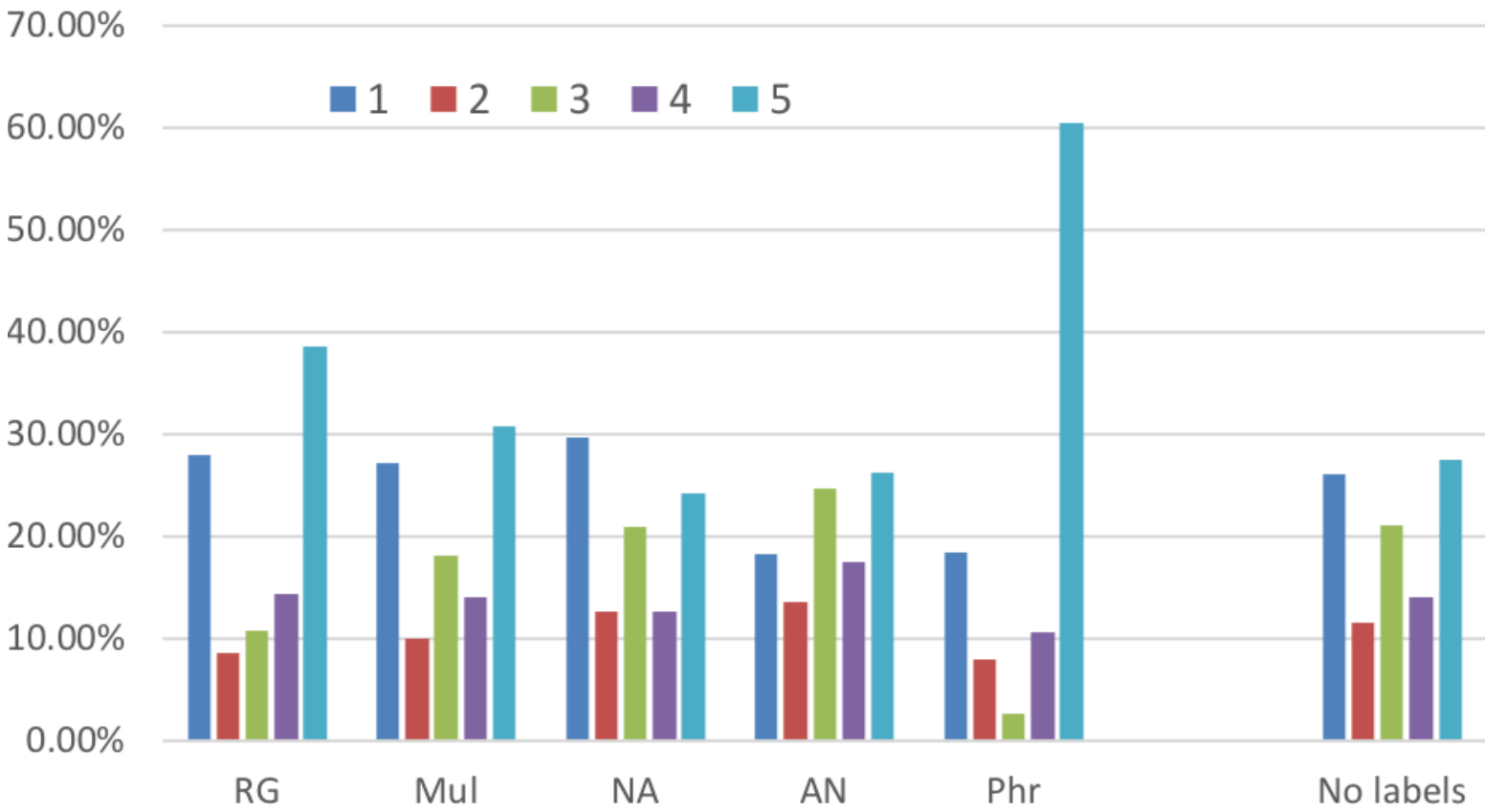}
 \caption{Within-label distribution of individual ratings of scenarios, grouped and normalized by qualitative scenario label (x-axis).}
\label{fig:cooccurrence}
\end{figure}

The scenarios were manually annotated with features of potential interest. After an initial inspection of the responses from Stage 2, features were selected if the authors suspected that they would identify scenarios likely to affect ratings in some significant way. The features, along with their abbreviations and criteria, are as follows:

\begin{itemize}
    \item \textbf{Phrase repeated (Phr)} - The scenario description repeats the provided phrase, or important key words from it, in such a way that points to a desired interpretation of the phrase. For example, if the phrase were ``serving the public good," the scenario ``the professor contributed to the public good by doing honest research" would be flagged. This does not include uses of the provided phrase or its key words if they are not used to describe the action. For example, given the phrase ``acting with the highest integrity", the scenario ``the professional launched an investigation into a colleague who is widely regarded to act with the highest integrity" would not be flagged, because the repeated phrase does not describe the action performed (the launching of an investigation).
    \item \textbf{No action by actor (NA)} - The scenario description does not describe an action performed specifically by the actor the rule is designed to cover, e.g. ``a teacher is guilty of abuse". If there is an action, it is either performed by someone else, the action is described using passive voice, or the actor is the object of the action. Non-actions such as ``the architect never swims" would be labeled. In contrast, ``the architect refuses to swim" is considered an action. Also note that the actor need not necessarily be an individual; in ``the architect's firm does X," the firm still falls under the code of ethics of architects. Even if it is strongly implied who the actor is, the actor needs to be explicitly stated: ``books are loaned to the public" in the `library' code of ethics would be labeled.
    \item \textbf{Action by non-actor (AN)} - An action is described that is either performed by someone other than the actor the rule is designed to cover, or an action is described using passive voice (``the teacher receives information"). This includes cases where the passive action is merely provided as context to the primary action, e.g. ``the teacher receives information about the student, and after receiving this info, performs action X". It's possible a scenario to have both this label and \textbf{NA}, if the only action described is not performed by the actor. 
    \item \textbf{Multiple actions by main actor (Mul)} - The scenario description describes multiple actions by the actor the rule is designed to cover. If additional actions are described but there is only one action performed by the primary actor, this label does not apply.
    \item \textbf{Reason given (RG)} - The scenario description provides a reason or intention behind the action, e.g. ``X does Y so that Z", ``X does Y in order to Z", etc. Must explicitly link to the reason in a causal way, e.g. ``the architect destroys old decrepit buildings and replaces them" would not be labeled; but ``the architect destroys old decrepit buildings so he can replace them" would. ``The professor uses mesh for his hammock" is not labeled because ``for his hammock" is more of a description rather than an intention.
\end{itemize}
Table \ref{tbl:scenarioLabels} lists these labels along with their frequencies, average ratings, and the average standard deviations of the ratings (as calculated within each scenario). Figure \ref{fig:cooccurrence} shows the distribution of individual ratings within scenarios with each label. Given each label $l$ and numerical rating value $n$, the number of times that a rating of $n$ was assigned to a scenario labeled with $l$ was divided by the total number of ratings of a scenario labeled with $l$. The resulting value is the y-axis in Figure \ref{fig:cooccurrence}. For comparison, the distribution of ratings within scenarios that received no labels is also shown. A few observations can be made:
\begin{itemize}
    \item  The \textbf{RG} and \textbf{Phr} labels seem to have the largest percentage of ratings at the extremes; respectively, they have 67\% and 79\% of their ratings as `1' or `5,' as opposed to 53\% for scenarios with no annotations. In \textbf{RG} cases, we suspect this is because explaining the intention behind an action makes it easier to empathize with an actor \cite{Dennett1989}. In \textbf{Phr} cases, repeating parts of the target phrase often appeared with explicit modifiers (e.g., if the phrase were ``behaving kindly," a scenario description might be ``the scientist treated his colleague unkindly," nakedly suggesting an intended interpretation).
    \item \textbf{Phr} had the lowest standard deviations of within-scenario ratings, suggesting that the repeating of a phrase within the scenario description does indeed encourage less diversity in interpretations. 
    \item \textbf{AN} had the lowest percentage of ratings at extremes (44\%). This may reflect confusion by the raters on which action to interpret. For example, in the scenario ``the doctor was not paid as much as her colleagues,'' it may be unclear whether which action the target phrase is being applied to: that of the unnamed individual or group which decides how much the doctor is paid, or some unstated action of the doctor that may be responsible for the pay disparity. 
\end{itemize}

\begin{figure}[ht]
\centering
\includegraphics[width=\columnwidth]{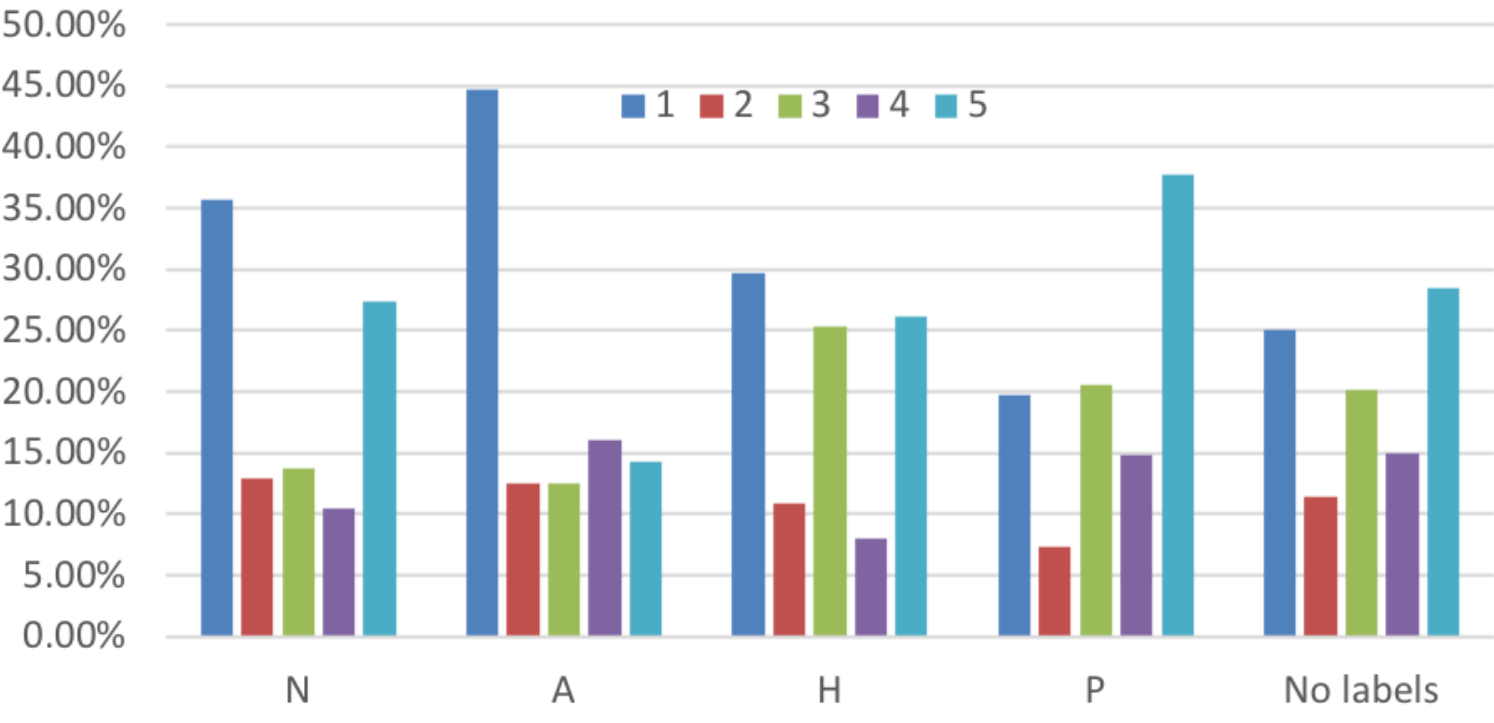}
 \caption{Within-label distribution of individual ratings, grouped and normalized by qualitative phrase label (x-axis).}
\label{fig:phrase_cooccurrence}
\end{figure}

\begin{center}
\begin{table}[]
\begin{tabular}{|c|c|c|c|}
\hline
\textbf{Label} & \textbf{Count} & \textbf{\begin{tabular}[c]{@{}c@{}}Average Rating\end{tabular}} & \textbf{\begin{tabular}[c]{@{}c@{}}Average Stdev\end{tabular}} \\ \hline
\textbf{N} & 11 & 2.808 & 0.915 \\ \hline
\textbf{A} & 3 & 2.614 & 0.971 \\ \hline
\textbf{H} & 5 & 2.853 & 0.704 \\ \hline
\textbf{P} & 8 & 3.333 & 0.861 \\ \hline
\textbf{No label} & 52 & 3.106 & 0.844 \\ \hline
\end{tabular}
\caption{Qualitative Labels Added to Phrases}
\label{tbl:phraseLabels}
\end{table}
\end{center}

\subsection{Phrase Labels}

In order to determine whether the phrases affected had some effect on the scenarios or their interpretations, we also annotated the phrases with the following labels:

\begin{itemize}
    \item \textbf{Negative actions (N)} - The actions which are typical instances of the phrase are generally, on the surface, unethical behaviors. E.g., `causing disrepute,' `making unwarranted statements,' `acting against public interest,' etc.
    \item \textbf{Absence actions (A)} - The phrase is primarily defined as an absence of an action, or acting against an action, rather than an action itself. E.g., `acting against public interest,' `not participating,' etc. Phrases such as ``engaging in misconduct'' were not annotated with this label, as `misconduct' is a distinct type of action rather than merely the absence of proper conduct.
    \item \textbf{Having (H)} - Phrase starts with the word `having.' 
    \item \textbf{Public (P)} - Phrase contains the word `public.' Such phrases seem particularly inviting of a broad variety of interpretations, since so many actions can be justified as being in service of the `public good.' 
\end{itemize}
The average ratings and standard deviations of ratings within-scenario for all scenarios created under each annotated label are listed in Table \ref{tbl:phraseLabels}. Figure \ref{fig:phrase_cooccurrence} shows the distribution of individual ratings for all ratings of scenarios under each label type. Given each phrase label $l$ and numerical rating value $n$, the number of times that a rating of $n$ was assigned to a phrase-scenario pair where the phrase was labeled with $l$ was divided by the total number of ratings assigned to a phrase-scenario where the phrase was labeled with $l$. The resulting value is the y-axis in Figure \ref{fig:phrase_cooccurrence}.

In Figure \ref{fig:phrase_cooccurrence} it is clear that phrases annotated with \textbf{N} and \textbf{A} have relatively high occurrences of `1' ratings. Recall that when the scenarios were created in Stage 2, approximately one-third of them were written with the intent to be interpreted as clear non-instances of their corresponding phrases. If the phrase labels \textbf{N} and \textbf{A} simply pick out scenarios that were intended to be clear non-instances, this would explain the high occurrence of `1' ratings. But this is not the case: phrases annotated with \textbf{N} or \textbf{A} were written with the intent to be clear non-instances 38.7\% of the time, as compared to 33.0\% 
for phrases without those labels. Instead, it may be the case that \textbf{N} and \textbf{A} phrases have a negative biasing effect on interpretations (one which sentiment-based automated interpretive reasoners should watch out for). Another possible explanation is that the additional cognitive load of interpreting a phrase describing the negation or absence of an action may have led to more diversity in interpretive arguments (indeed, the within-scenario standard deviation of ratings was higher for phrases \textbf{N} and \textbf{A}).

Our initial suspicion was that \textbf{H} phrases would produce scenarios which were unclear about actions in some way. For example, the phrase ``having substantial information'' describes a state or property rather than an action, and thus scenarios generated when prompted with such phrases may be more likely to describe actions by non-actors (\textbf{AN}), or perhaps fail to describe actions at all (\textbf{NA}). \textbf{H} phrases did produce the lowest standard deviation of ratings within-scenarios, and the highest number of `3' scores, when compared to other phrase labels. However, when determining the correlations between each phrase label and scenario label, no significant correlations were found.

%% file: individual.tex
\section{Individual Case Analysis}

Analysis of individual cases is a useful research methodology for exploratory, descriptive, or explanatory works \cite{Yin1981} as it deals with information in a complex, holistic, process-oriented, and particularistic way that is reflective of reality \cite{Wilson1979}. In this section, we analyze individual responses by trying to answer three questions: (1) What kind of reasoning could have been used to generate this response? (2) Can current state-of-the-art AI carry out this kind of reasoning? And (3) if not, which capabilities does AI need, and which questions must AI-related researchers answer, before it can?


\subsection{Details of Questionable Relevance} Perhaps because the scenario descriptions were required to be no more than two sentences, raters in Stage 3 may have assumed that any details included in the scenario description must be important. Any details which would normally be of questionable relevance therefore may have a heightened effect in this task, making their effect easier to identify. E.g.:

\nicebox{1}{
    \textbf{Rule:} To perform all professional work in a manner that merits full confidence and trust; to be conservative in estimates, reports, and testimony, especially if these are related to the promotion of a business enterprise or the protection of the public interest.\\
    \textbf{Phrase:} Protecting the public interest\\
    \textbf{Scenario:} The chemist filled out his tax report.\\
    \textbf{Ratings / Justifications:}
    \begin{enumerate}
        \item (Rating 1) The tax report is irrelevant to the public interest.
        \item (Rating 5) Filing taxes is a clear example of protecting the public interest because taxes are considered public funds.
    \end{enumerate}
}
The action of filling out a tax report appears on the surface to be completely outside of the domain described by the phrase or the rule and its associated profession (chemistry). Statistically calculating pre-existing associations between this action and domain, as might be done by many currently popular machine learning approaches, would likely result in low values. But interpretive reasoning is more than reasoning about pre-existing associations; it also requires reasoning about possible connections which can be supported by interpretive arguments. The first rater was not able to find such a connection, whereas the second rater was: taxes are a type of public funds, public funds are vital to the public interest; therefore, filling out taxes protects the public interest. 

A similar example comes from the code of ethics for Massage Therapists:

\nicebox{2}{
    \textbf{Rule:} The Massage Therapist shall project a professional image and uphold the highest standards of professionalism.\\
    \textbf{Phrase:} Projecting a professional image\\
    \textbf{Scenario:} The massage therapist brushes her teeth every morning.\\
    \textbf{Ratings / Justifications:}
    \begin{enumerate}
        \item (Rating 5) A simple [sic] but an example of maintaining a professional image by hygiene.
        \item (Rating 4) While trivial, the therapist brushing her teeth is an example of projecting a professional image. It is only a ``somewhat clear" example because it is trivial and probably not what the rule is intended to cover.
        \item (Rating 5) Given that massage is the health and wellness industry, clean teeth do help project a professional image.
        \item (Rating 1) Scenario act should be performed anyway.
    \end{enumerate}
}
Raters 1 to 3 note that the act of brushing one's teeth seems trivial, but all three conclude it is not entirely irrelevant to projecting a professional image. Rater 4 disagrees, specifically because the act described in the scenario is one that is already a normal everyday expectation of a massage therapist. But which conclusion is more warranted from the premise that brushing one's teeth is an everyday expected task: that brushing one's teeth is an instance of projecting a professional image that only minimally contributes to it, or that it is \textit{not} an instance at all?

The ability to answer such questions might be expected of an agent capable of evaluating interpretive arguments, whether artificial or human. It might be the case that an artificial interpretive reasoner, being necessarily bound by time limitations, might be hard-coded with a preference for interpretations that maximize practical consequence. In such a case, what sort of algorithm would allow a massage therapy-bot to calculate the practical consequence of brushing its teeth (or the equivalent act for robots)?


\subsection{Omission of Details} As opposed to introducing irrelevant or misleading details, another strategy that invited disagreeing interpretations was the omission of potentially relevant details in scenario descriptions. For example: 

\nicebox{7}{
    \textbf{Rule:} Employers should provide [chemical professionals] with a safe, congenial working environment, fair compensation, opportunities for advancement, and properly acknowledge their scientific contributions.\\
    \textbf{Phrase:} Providing fair compensation\\
    \textbf{Scenario:} The chemical professional receives higher wage compared to colleagues in similar roles.\\
    \textbf{Ratings / Justifications:}
    \begin{enumerate}
        \item (Rating 1) People in similar roles should receive similar wage compensation
        \item (Rating 5) Better than market compensation is more than fair.
        \item (Rating 1) Since one person is receiving higher wages than colleagues in a comparable position, this is inconsistent with the concept of providing fair compensation because similarly qualified individuals are not being treated similarly.
    \end{enumerate}
}
In this case, the reasons for the wage disparity are not specified, only that the chemical professional and colleagues are in ``similar roles.'' Raters 1 and 3 interpret the action of `providing fair compensation' as encompassing fair treatment of all workers, whereas the second rater interprets a higher salary than colleagues as being ``more than fair" for the individual receiving the higher salary. 

In cases where the scenario description is extremely short and sparse on details, raters often had to draw on their own background knowledge to fill in the details:

\nicebox{6}{
    \textbf{Rule:} Model the trustworthiness of God in leadership to encourage and develop trustworthiness in others. Use power and influence prudently and humbly. Foster loyalty. Demonstrate a commitment to the well-being of the entire congregation.\\
    \textbf{Phrase:} Committing to the entire congregation's well-being\\
    \textbf{Scenario:} The pastor cleans the altar.\\
    \textbf{Ratings / Justifications:}
    \begin{enumerate}
        \item (Rating 5) While it appears to be a menial task, the pastor's cleaning the altar himself is a demonstration of commitment to the well-being of his or her congregation.
        \item (Rating 1) The altar is only used by those that officiate at the service.  Most of the congregation is in the audience and are not affected by the altar's cleanliness.
    \end{enumerate}
}
In both Stages 2 and 3, mTurk participants were shown the profession type whose code of ethics the rule was from. In the above example, the profession was ``pastor,'' and the interpretations here show that the raters drew heavily from their own preconceptions of the normal operations of a pastor's job. In the first case, the rater noticed that the pastor cleaned the altar himself (presumably as opposed to having another clean it). In the second, the rater points out that the altar is not normally used by most of the congregation. Although these two interpretations have entirely opposite numerical ratings, they both start by identifying the significance of the action of cleaning the altar within the context of the pastor's normal duties. Automated interpretive reasoners, then, must rely on a significant amount of background knowledge \cite{Licato2019b}.

\subsection{Contrarian Arguments} A third common cause of disagreement in interpretive arguments may best be described as a desire to be contrarian, i.e., by providing a non-standard interpretation that, whether intentionally or not, goes against the majority opinion (recall the example argument $I_1$). Generating contrarian arguments requires the ability to anticipate the common or expected reaction to a scenario, and then to argue for the opposite reaction (both of which are abilities seemingly beyond the state-of-the-art of current AI). Reasons for such a behavior may vary---perhaps the participant is bored, wants to exercise creative reasoning, may want to be intentionally malicious, etc.---but in the context of group decision-making, the occasional desire to stand against majority opinions (sometimes called ``playing devil's advocate'') can prove beneficial for the group overall \cite{Licato2018c}.

Intentional contrarianism can have real world consequences. A malicious individual might try to convince an artificial reasoner that some non-standard interpretation of a phrase in an ethical rule is correct. It is therefore worthwhile for automated interpretive reasoning research to understand how to detect contrarian interpretive arguments, or at least determine whether a typical human audience would consider them contrarian. Consider the following example:


\nicebox{3}{
    \textbf{Rule:} A veterinarian shall uphold the standards of professionalism, be honest in all professional interactions, and report veterinarians who are deficient in character or competence to the appropriate entities.\\
    \textbf{Phrase:} Having deficient character\\
    \textbf{Scenario:} The veterinarian beats a dog in his kennel who is being inappropriately loud or violent.\\
    \textbf{Ratings / Justifications:}
    \begin{enumerate}
        \item (Rating 1) A veterinarian beating a dog seems to be the definition of unprofessional, clearly shows a character deficiency but has nothing to do with the requirement to ``report veterinarians who are deficient in character or competence to the appropriate entities." If one element of a conjunction is false then so is the conjunction.
        \item (Remaining 3 ratings were `5')
    \end{enumerate}
}
The first rater's interpretation interprets wording in the rule outside of the phrase to limit its scope (although it references a logical conjunction, the source of which is difficult to reconstruct). Such a practice is not uncommon in legal reasoning, but was uncommon in this dataset, likely because the questions in Stage 3 asked participants to use the phrase to determine interpretations while merely \textit{considering} the rule. Rater 1 may not intentionally have been acting in an intentionally contrarian role, but nevertheless plays the role. In any case, the rater likely recognized that their interpretation was non-standard, as they prefaced their argument with a sort of disclaimer statement acknowledging what they anticipate the expected interpretation to be.